\documentclass[letterpaper]{article} % DO NOT CHANGE THIS
\usepackage{aaai2027}  % DO NOT CHANGE THIS
\usepackage[hyphens]{url}  % DO NOT CHANGE THIS
\usepackage{graphicx} % DO NOT CHANGE THIS
\usepackage{natbib}  % DO NOT CHANGE THIS AND DO NOT ADD ANY OPTIONS TO IT
\usepackage{caption} % DO NOT CHANGE THIS AND DO NOT ADD ANY OPTIONS TO IT
\usepackage{algorithm}
\usepackage{algorithmic}
\usepackage{amsmath}
\usepackage{multirow} 
\usepackage{makecell}
\usepackage{subcaption}
\usepackage{booktabs} 
\usepackage{xcolor}
\usepackage{listings}
\usepackage{mdframed}

\newenvironment{promptblock}[1]{
\begin{mdframed}[
  linewidth=0.6pt,
  linecolor=black!55,
  backgroundcolor=white,
  roundcorner=2pt,
  innertopmargin=0pt,
  innerbottommargin=4pt,
  innerleftmargin=5pt,
  innerrightmargin=5pt,
  skipabove=4pt,
  skipbelow=4pt
]
\noindent\colorbox{black!55}{
  \parbox{\dimexpr\linewidth-2\fboxsep\relax}{
    \color{white}\bfseries\small #1
  }
}
\vspace{2pt}
}{
\end{mdframed}
}

\usepackage{newfloat}
\usepackage{listings}
\DeclareCaptionStyle{ruled}{labelfont=normalfont,labelsep=colon,strut=off} % DO NOT CHANGE THIS
\floatstyle{ruled}
\newfloat{listing}{tb}{lst}{}
\floatname{listing}{Listing}

\usepackage{booktabs}

\title{FTA-Mem: Fact-Time-Affect Anchored Memory for Low-Density Long-Term Dialogue}
\author{
     \textbf{Chang Liu\textsuperscript{1}},
     \textbf{Junyi Zhao\textsuperscript{1}},
 \textbf{Shuyi Zhang\textsuperscript{1}},
 \textbf{Changsheng Ma\textsuperscript{1}},
 \textbf{Yongfeng Tao\textsuperscript{1}},\\
 \textbf{Minqiang Yang\textsuperscript{1}},
 \textbf{Bin Hu\textsuperscript{1}},
}
\affiliations{
    \textsuperscript{1}School of Information Science and Engineering, Lanzhou University,
}

\begin{document}

\maketitle

\begin{abstract}
Long-term emotional-support agents require memory mechanisms for personalized understanding across sessions. However, emotional-support dialogue is often low-density: turns are incomplete, evidence is scattered, and user states evolve over time. Existing memory methods usually rely on fixed units, such as session summaries or turn-level notes, which may lose details or introduce redundant noise.
We propose FTA-Mem, a structured memory framework for low-density long-term dialogue. FTA-Mem uses \textbf{B}oundary-preserving \textbf{W}indow \textbf{S}egmentation (\textbf{BWS}) to form coherent situation fragments, and constructs \textbf{F}act-\textbf{T}ime-\textbf{A}ffect Memory Units (\textbf{FTA} Units) that jointly encode factual content, temporal grounding, and affective context. Retrieved units are then synthesized into structured context for answer generation.
Experiments on ES-MemEval and LoCoMo show that FTA-Mem improves overall long-term memory question answering across benchmarks with different information-density characteristics. On ES-MemEval, FTA-Mem achieves 0.3871 F1 and 0.6668 BERTScore. Further analysis shows that situation-level FTA construction better balances evidence preservation and construction cost than coarse session-level or overly fine-grained turn-pair construction, providing an effective granularity trade-off for long-term dialogue memory.
\end{abstract}


\section{Introduction}

Large language models (LLMs) have shown substantial potential as conversational agents and are increasingly used in customer support, emotional support, and mental health services~\citep{chen2024survey, zou2026llm}. Although these models can generate fluent and empathetic responses in short-term interactions, psychological and emotional support is inherently longitudinal~\citep{schaie1983longitudinal}. Reliable long-term support requires an agent to maintain personalized understanding of users' experiences, emotional trajectories, and prior support interactions across sessions. Building robust long-term memory systems has therefore become a central challenge for long-term conversational agents~\citep{hatalis2023memory, maharana2024evaluating, wu2026memory}.

However, long-term emotional-support dialogue poses a distinct memory challenge: it is often low-density. A single turn rarely contains a complete, self-contained fact; useful evidence may be sparse, indirect, temporally ambiguous, or distributed across multiple sessions~\citep{chen2026memeval}. Moreover, the meaning of an event often depends not only on what happened, but also on when the information remains valid and how the user appraises it. Therefore, simply retrieving more history or compressing longer contexts does not necessarily yield reliable memory.

Many long-term memory systems for LLM agents follow a retrieval-augmented paradigm: past interactions are converted into retrievable memory units and later selected as context for generation. Some systems keep memory close to the original dialogue, retrieving raw passages or compact summaries and profiles~\citep{zhong2024memorybank}. Others introduce managed memory stores or agentic update mechanisms to support persistence and revision~\citep{packer2024memgptllmsoperatingsystems,kang2025memory}. More recent work further organizes memory with graph links or event segmentation, connecting related semantic units or merging adjacent turns into event memories~\citep{xu2026mem,ke2025flexibly,zou2026esmemeventsegmentationbasedmemory}. These designs improve storage, updating, or retrieval from different angles, but their effectiveness still depends on the quality and granularity of the underlying memory units.

This raises two basic but underexplored questions: \textbf{at what granularity should low-density long-term dialogue memories be constructed, and what should count as a retrievable memory unit?} Coarse units such as session summaries are compact, but may omit QA-sensitive details such as temporal updates, unresolved plans, relationship changes, and affective cues. Fine-grained units such as turns or turn pairs preserve local evidence, but often split a coherent user event or state into redundant fragments.  Recent event-segmentation and event-centric memory methods improve semantic integrity by grouping adjacent turns into segments or using LLMs to extract coarse event graphs~\citep{wu2025sgmemsentencegraphmemory, zou2026esmemeventsegmentationbasedmemory,hu-etal-2026-memory}. However, low-density psychological dialogue requires organizing distributed factual, temporal, and affective cues within memory units, rather than only forming coarse event groups.

To address this issue, we propose FTA-Mem, a structured memory framework for low-density long-term dialogue.  FTA-Mem constructs situation-level Fact-Time-Affect Memory Units. FTA-Mem first preserves broader contextual continuity with Boundary-preserving Window Segmentation (BWS), and then decomposes each coherent fragment into compact memory units. It also maintains units around segment boundaries by carrying unresolved units across fragments, fusing adjacent-fragment units before persistent ID assignment, and linking finalized units to related historical memories. Each unit jointly grounds factual evidence, temporal validity, and affective interpretation. During answer generation, retrieved FTA units are synthesized into structured context packets rather than passed as a flat list of retrieved passages.

Our contributions are:
\begin{itemize}
    \item We introduce FTA-Mem, a structured memory framework that represents low-density long-term dialogue with situation-level Fact-Time-Affect memory units.
    \item We design a boundary-preserving construction pipeline that preserves contextual continuity through unit carryover, adjacent-fragment fusion, and temporal-link maintenance.
    \item We evaluate FTA-Mem on ES-MemEval~\citep{chen2026memeval} and LoCoMo~\citep{maharana2024evaluating}, showing improved question-answering performance and a better granularity trade-off than session-level and turn-pair memory construction.
\end{itemize}

\section{Related Work}

\subsection{Psychological Counseling Agents}

Early LLM-based counseling agents mainly focus on response generation in single-session or short-term interactions, such as CBT-LLM~\citep{na2024cbt} for CBT-style counseling and HealMe~\citep{xiao2024healme} for short multi-turn emotional support. These systems improve counseling responses but do not explicitly model long-term memory~\citep{li2025hello, wang2026psychological}.
Recent work has begun to move toward multi-session psychological counseling. MusPsy~\citep{wang2026psychological} synthesizes multi-session counseling cases from user profiles and psychological scenarios. TheraMind~\citep{hu2026theramind} and PsychAgent~\citep{yang2026psychagentexperiencedrivenlifelonglearning} further target continuous counseling settings through strategy planning, client profiling, memory-augmented continuity, and skill evolution. These systems highlight the importance of long-term support, but their memory components are typically organized around session summaries, client profiles, planning states, or therapeutic skills. In contrast, FTA-Mem focuses on the representation of retrievable memory units themselves, preserving factual evidence, temporal validity, and affective state in low-density emotional-support dialogue.

\subsection{Long-Term Memory for LLM Agents}

Retrieval-augmented generation (RAG) mitigates the context limitation of LLMs by retrieving external information~\citep{fan2024survey, arslan2024survey}. Existing long-term memory systems for LLM agents mainly differ in how they define and maintain memory units. GraphRAG demonstrates the utility of graph-based organization for memory retrieval~\citep{edge2025localglobalgraphrag}; MemoryBank stores dialogue history through summaries~\citep{zhong2024memorybank}; MemGPT and MemoryOS introduce hierarchical memory management~\citep{packer2024memgptllmsoperatingsystems,kang2025memory}; A-MEM builds agentic memory notes with links and updates~\citep{xu2026mem}; FraCom extracts turn-derived semantic units and connects them with graph structures~\citep{ke2025flexibly}; ES-Mem segments adjacent turns into coherent event memories~\citep{zou2026esmemeventsegmentationbasedmemory}; and CompassMem organizes dialogue into event-centric memory maps~\citep{hu-etal-2026-memory}.  However, their underlying units are often still raw dialogue turns, summaries, or turn-derived semantic units.
This leaves the granularity of retrievable memory underexplored in low-density long-term dialogue. Coarse units, such as session summaries, are compact but may lose QA-sensitive details; fine-grained turn or turn-pair units preserve local evidence but may split coherent situations into redundant fragments. FTA-Mem therefore focuses on the retrievable unit itself, constructing situation-level Fact-Time-Affect units that preserve factual evidence, temporal grounding, and affective state before retrieval, linking, and context synthesis.

\begin{figure*}[t]
\centering
\includegraphics[width=0.98\textwidth]{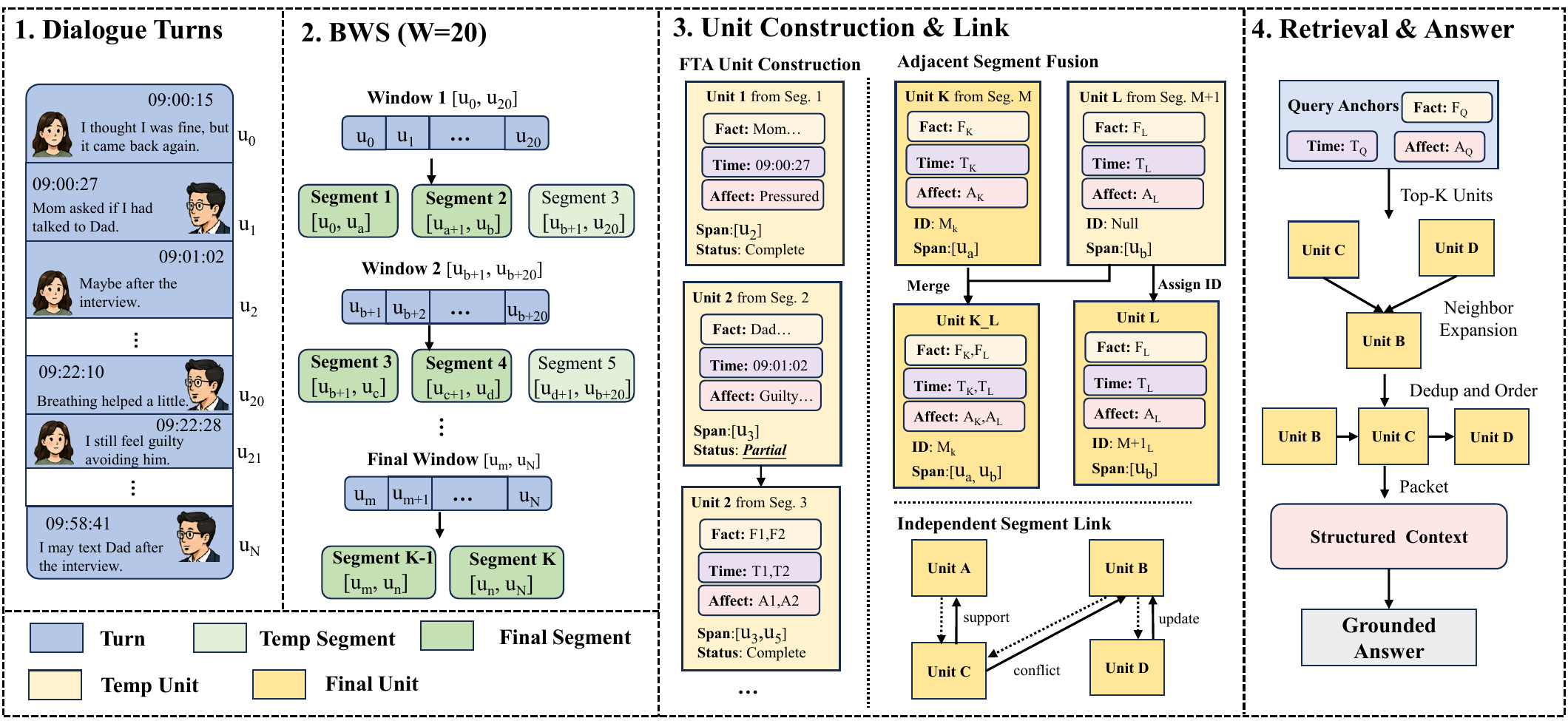} 
\caption{Overview of FTA-Mem. BWS segments dialogue turns into boundary-preserving fragments, which are converted into Fact-Time-Affect units. Partial units can be carried forward, adjacent units can be fused before ID assignment, and finalized units are linked, retrieved, deduplicated, and organized into structured context for grounded answer generation.}
\label{fig:framework}
\end{figure*}

\section{Methodology}

\subsection{Problem Definition}

We consider a long-term dialogue case $C$ consisting of multiple sessions $C=\{S_1,S_2,\ldots,S_N\}$ ,
where each session is $S_i=\{u_{i,1},u_{i,2},\ldots,u_{i,T_i}\}$. Each turn $u_{i,t}$ contains the speaker role, timestamp, and utterance text. Given a user query $q \in Q$, the goal is to generate an answer $a$ grounded in the case history.

Directly conditioning on $C$ is infeasible for long cases and unreliable for low-density dialogue, where useful evidence may be sparse, implicit, or distributed across sessions. FTA-Mem therefore introduces an external structured memory $M$ as an intermediate representation. At inference time, the query retrieves a subset of relevant memory units $M_{sub} \subset M$, and the answer is generated as:
\begin{equation}
a^*=\arg\max_a p_\theta(a \mid q, M_{sub}).
\end{equation}
Thus, the central problem becomes how to construct memory units that are compact enough for retrieval, yet expressive enough to preserve factual evidence, temporal validity, and affective context. The overall pipeline of FTA-Mem is illustrated in Figure~\ref{fig:framework}.

\subsection{Boundary-preserving Window Segmentation}

Low-density emotional-support dialogue often requires surrounding context to form a complete memory. A single turn may be incomplete, while an entire session may be too long and heterogeneous. We therefore introduce \textbf{Boundary-preserving Window Segmentation} (BWS), inspired by event segmentation theory~\citep{zacks2007event}, to form coherent situation fragments before memory construction.

For each session $S_i$, we first attach a numeric turn index to every turn. Given a window size $W$, BWS applies an LLM-based boundary detector $g_\phi$ to a consecutive window:
\begin{equation}
g_\phi(S_i[s:s+W-1]) \rightarrow \{(b_j,e_j,r_j)\}_{j=1}^{K},
\end{equation}
where $(b_j,e_j)$ denotes a continuous turn span and $r_j$ denotes the boundary cue, such as a topic change or emotional shift.

To avoid context truncation caused by window boundaries, BWS does not simply move forward by a fixed stride. After segmenting the current window, it repositions the next window to the start of the last detected segment, i.e., $s \leftarrow b_K$. BWS finalizes completed segments before the last detected segment, keeps the last segment provisional for the next window, and enforces a minimum forward step when the detector returns a single full-window segment. If the current window reaches the end of the session, BWS finalizes the detected segments and terminates. This allows boundary-uncertain content near the window tail to be reprocessed with following context.

The output of BWS is a sequence of contiguous fragments $F_i=\{f_{i,1},f_{i,2},\ldots,f_{i,L_i}\}$ for each session $S_i$. Each fragment is then used as the evidence span for constructing one or more memory units.

\subsection{Memory Unit Construction}

FTA-Mem is inspired by Tulving's distinction between episodic and semantic memory~\citep{tulving1972episodic}. At the episodic level, it constructs evidence-grounded Fact-Time-Affect units from dialogue fragments; at the longitudinal level, it can maintain semantic user memory and support-experience memory for consistency.

\subsubsection{Fact-Time-Affect Memory Unit}

FTA-Mem represents long-term dialogue at the situation level. A situation may correspond to a user event, an evolving state, or a support interaction. Following situation models~\citep{zwaan1998situation}, a useful memory unit should preserve what happened, when it is valid, and how it is affectively situated.

The basic memory node is a \textbf{Fact-Time-Affect Memory Unit}, denoted as $m=\langle x^F,x^T,x^A,e,o\rangle$. Here, $x^F$ is the fact anchor, $x^T$ is the time anchor, $x^A$ is the affect anchor, $e$ is the evidence pointer, and $o$ is the unit-construction status. The fact anchor stores the factual claim, situation type, participants, and summary; the time anchor stores dialogue time, event time, temporal orientation, and situation completion status; and the affect anchor stores broader subjective context, including emotion, intention, and relation cues.

Given a situation fragment $f_{i,j}$ and unresolved units $I_{i,j}$, an LLM-based extractor constructs candidate FTA units:
\begin{equation}
\{m_{i,j}^{k}\}_{k=1}^{K_{i,j}}
=
\mathrm{Ext}_{\theta}(f_{i,j}, I_{i,j}).
\end{equation}
The unit-construction status $o$ indicates whether the memory unit is complete, partial, pending, or unknown. To reduce segmentation-induced incompleteness, unresolved partial units are carried to the next fragment:
\begin{equation}
I_{i,j+1} =
\{m \in M_{\leq i,j} \mid o(m)=\mathrm{partial}\}.
\end{equation}
Finalized units are assigned persistent memory IDs and inserted into the memory store, while unresolved partial units remain construction context rather than independently retrievable memories.

\subsubsection{Unit Consolidation and Temporal-Link Maintenance}

FTA-Mem uses a two-level maintenance mechanism to reduce segmentation-induced redundancy while preserving longitudinal consistency. The first level locally consolidates partial units across adjacent fragments, and the second level maintains temporal links among distributed memories.

After the extractor produces candidate units for fragment $f_{i,j}$, FTA-Mem first checks whether each candidate should be fused with an unresolved unit carried from adjacent fragments. This step targets cases where a situation near a fragment boundary is partially extracted in one fragment and completed in the next. If a new candidate $\tilde{m}$ is compatible with a previous unresolved unit $m_p \in I_{i,j}$ and can complete or refine it, FTA-Mem performs local fusion:
\begin{equation}
m_p \leftarrow \mathrm{Fuse}(m_p,\tilde{m}).
\end{equation}
The fusion combines their factual anchors, temporal anchors, affective context, and evidence pointers. The new candidate is not assigned an independent memory ID. If no local fusion is triggered, the candidate is finalized, assigned a persistent memory ID, and inserted into the memory store.

FTA-Mem then links each newly stored unit to related historical units. Candidate neighbors are retrieved from the existing memory store by embedding similarity, and a relation classifier labels each pair as same-situation, update, contradiction, follow-up, support, or unknown. Non-unknown relations are added to the memory graph:
\begin{equation}
\mathcal{E}_r \leftarrow \mathcal{E}_r \cup \{(m,m',r)\}.
\end{equation}
The graph supports bidirectional access during retrieval, while each relation label preserves its direction from the new unit to the historical unit. These links update relation labels, temporal validity, lifecycle status, and reverse pointers for later neighbor expansion.
In this way, FTA-Mem can track whether a prior situation has been completed, revised, contradicted, or supported over time.

\subsubsection{Longitudinal Memory Maintenance}

In addition to episodic FTA units, FTA-Mem maintains auxiliary longitudinal memory after each session, including user semantic memory $\mathcal{U}$ and support-experience memory $\mathcal{H}$. Given episodic units $\mathcal{M}_i$ from session $S_i$, the memories are updated as:
\begin{equation}
\begin{aligned}
\mathcal{U}_i &= \mathrm{Update}_U(\mathcal{U}_{i-1}, \mathcal{M}_i), \\
\mathcal{H}_i &= \mathrm{Update}_H(\mathcal{H}_{i-1}, \mathcal{M}_i^{sup}).
\end{aligned}
\end{equation}
They provide auxiliary context, while episodic FTA units remain the primary evidence source.

\subsection{Retrieval and Context Synthesis}

At inference time, FTA-Mem retrieves relevant FTA units and synthesizes them into a structured memory packet for answer generation.

\subsubsection{Query Rewriting and Unit Retrieval}

Given an input query $q$, FTA-Mem rewrites it into an evidence-oriented retrieval query $q'$. The rewrite preserves explicit entities, temporal expressions, relations, and task intent, so that the query better matches the structured memory store rather than broad semantic summaries.

Each memory unit $m$ is represented by a structured retrieval text $\rho(m)$, constructed from its factual anchor, temporal anchor, affective anchor, and evidence metadata. FTA-Mem scores each unit as:
\begin{equation}
\label{eq:retrieval_score}
R(q',m)=
\lambda s_{\mathrm{emb}}(q',\rho(m))
+(1-\lambda)c(q',m),
\end{equation}
where $s_{\mathrm{emb}}$ is embedding similarity, $c(q',m)$ is a structured cue score over time, situation type, participants, and affective context, and $\lambda$ balances semantic similarity and structured cue matching.

The primary episodic evidence set is selected by thresholded top-$K$ retrieval:
\begin{equation}
M_{sub}=\mathrm{TopK}_{K}\{m\in M \mid R(q',m)\ge \tau\}.
\end{equation}
Here, $K$ controls the retrieval budget and $\tau$ filters weakly related memories.

\subsubsection{Linked Neighbor Expansion and Source Grounding}

The initially retrieved units may have later updates, contradictions, or follow-up evidence. FTA-Mem therefore expands a small number of linked neighbors:
\begin{equation}
L_{sub}=\mathrm{Expand}(M_{sub},\mathcal{E}_r,B),
\end{equation}
where $B$ limits neighbors per retrieved unit. The retrieved units and neighbors are deduplicated by persistent memory ID and ordered chronologically before context synthesis. These neighbors provide relation-aware support, while $M_{sub}$ remains the primary evidence set.

Each retrieved unit keeps an evidence pointer $e(m)$ to its original dialogue span. During context synthesis, FTA-Mem preserves these source pointers together with the unit content, so that the answer model can ground its response in the original conversation rather than only in abstracted memory text. In addition, FTA-Mem retrieves relevant user semantic memory $U_{sub}$ and support-experience memory $H_{sub}$, and groups them as auxiliary longitudinal memory \(A_{sub}=U_{sub}\cup H_{sub}\).

\subsubsection{Structured Context Synthesis}

FTA-Mem synthesizes the retrieved information into a structured memory packet:
\begin{equation}
P_q=\mathrm{Syn}(M_{sub},L_{sub},A_{sub}).
\end{equation}
The packet separates primary FTA units, linked neighbor units, source dialogue spans, and auxiliary longitudinal memory. This structured format prevents the answer model from treating all retrieved content as a flat passage list.

The final answer is generated as:
\begin{equation}
a^*=\arg\max_a p_\theta(a \mid q,P_q).
\end{equation}
During generation, $M_{sub}$ is treated as primary evidence, $L_{sub}$ as relation-aware supporting context, and $A_{sub}$ as auxiliary context.

\begin{table}[t]
\centering
{\small
\begin{tabular}{lcc}
\toprule
\textbf{Metric} & \textbf{ES-MemEval} & \textbf{LoCoMo} \\
\midrule
Turn length & 18.56 & \textbf{22.69} \\
Factual anchors / turn & 0.80 & \textbf{1.11} \\
Entity anchors / turn & 0.52 & \textbf{1.24} \\
Temporal anchors / turn & 0.11 & \textbf{0.22} \\
Support-process anchors / turn & \textbf{0.91} & 0.50 \\
Implicitness (1--3) & \textbf{1.91} & 1.58 \\
Low-info turns & \textbf{7.22\%} & 5.58\% \\
\bottomrule
\end{tabular}}
\caption{Turn-level information-density statistics of ES-MemEval and LoCoMo. Values are averaged over turns.}
\label{tab:density_analysis}
\end{table}

% \midrule

% \multirowcell{5}{Qwen3-4B}
% & MemoryBank & 24.92 & 29.73 & 21.76 & 45.44 & 24.83 & 29.02 & 54.97 & 63.47 & 55.53 & 51.06 & 65.23 & 58.25 & 0.964 & 4.0 \\
% & MemGPT     & \textbf{39.01} & 33.01 & 28.57 & 23.25 & 25.69 & 30.12 & \textbf{60.31} & \textbf{65.43} & 61.33 & 44.39 & \textbf{65.72} & 59.77 & 1.015 & 2.5 \\
% & MemoryOS   & 24.86 & 27.18 & 19.99 & \textbf{64.88} & 24.21 & \textbf{31.59} & 53.35 & 61.03 & 53.04 & \textbf{66.36} & 64.46 & 59.58 & 0.889 & 3.9 \\
% & A-Mem      & 31.78 & 33.20 & 25.14 & 40.78 & 26.02 & 31.23 & 57.59 & 65.17 & 57.50 & 51.54 & 65.62 & 59.70 & \textbf{1.018} & 2.4 \\
% & FTA-Mem    & 30.65 & \textbf{33.33} & \textbf{30.94} & 33.41 & \textbf{26.51} & 30.86 & 57.17 & 64.88 & \textbf{63.89} & 53.47 & 65.41 & \textbf{61.05} & \textbf{1.018} & \textbf{2.2} \\

\begin{table*}[t]
\centering
{\small
\setlength{\tabcolsep}{2.6pt}
\renewcommand{\arraystretch}{1.08}
\begin{tabular}{c l|rrrrrr|rrrrrr|rr}
\toprule
\multirow{2}{*}{\textbf{Backbone}}
& \multirow{2}{*}{\textbf{Method}}
& \multicolumn{6}{c|}{\textbf{F1 Score (\%) $\uparrow$}}
& \multicolumn{6}{c|}{\textbf{BERTScore (\%) $\uparrow$}}
& \multirow{2}{*}{\textbf{Judge $\uparrow$}}
& \multirow{2}{*}{\textbf{Avg.R $\downarrow$}} \\
& & IE & TR & CD & Abs & UM & All
& IE & TR & CD & Abs & UM & All
& & \\

\midrule
\multirowcell{6}{Qwen3-8B}
& MemoryBank & 23.13 & 29.76 & 25.65 & 33.29 & 24.55 & 27.08 & 53.50 & 64.45 & 59.95 & 49.37 & 65.31 & 58.66 & 1.036 & 4.3 \\
& MemGPT     & \underline{44.15} & \underline{32.77} & 25.14 & 29.54 & \underline{25.64} & \underline{31.69} & \underline{65.56} & \textbf{65.28} & 58.01 & 49.26 & \textbf{65.93} & \underline{61.19} & \underline{1.107} & \underline{2.8} \\
& MemoryOS   & 21.41 & 25.79 & 22.35 & \textbf{59.96} & 22.29 & 29.70 & 52.35 & 60.83 & 56.73 & \underline{67.48} & 63.87 & 60.09 & 0.917 & 4.9 \\
& A-Mem      & 27.96 & 31.66 & \underline{26.60} & 39.52 & 25.22 & 29.97 & 56.18 & \underline{65.22} & \underline{60.12} & 53.46 & \underline{65.55} & 60.23 & 1.090 & 3.0 \\
& CompassMem & 22.06 & 28.85 & 22.06 & 56.52 & 24.32 & 30.20 & 54.09 & 62.35 & 55.69 & 66.98 & 64.70 & 60.67 & 1.024 & 4.4 \\
& FTA-Mem    & \textbf{44.46} & \textbf{33.42} & \textbf{34.38} & \underline{56.55} & \textbf{26.35} & \textbf{38.71} & \textbf{67.80} & 64.40 & \textbf{65.62} & \textbf{70.26} & 65.54 & \textbf{66.68} & \textbf{1.119} & \textbf{1.5} \\

\midrule
\multirowcell{6}{Qwen3-14B}
& MemoryBank & 24.78 & 30.77 & 25.61 & 36.76 & 26.66 & 28.72 & 54.28 & 64.43 & 59.44 & 42.82 & 66.44 & 57.78 & 1.097 & 3.2 \\
& MemGPT     & \textbf{38.07} & \underline{33.96} & 21.85 & 40.47 & \underline{27.24} & \underline{32.33} & \underline{61.40} & \underline{66.34} & 54.43 & 44.81 & \underline{66.84} & 59.21 & 1.088 & \textbf{2.4} \\
& MemoryOS   & 21.51 & 28.46 & 21.91 & 30.83 & 24.91 & 25.40 & 51.53 & 62.07 & 55.78 & 41.39 & 65.14 & 55.48 & 0.952 & 4.4 \\
& A-Mem      & 30.30 & \textbf{34.58} & \underline{30.73} & 11.23 & \textbf{28.34} & 27.32 & 56.67 & \textbf{66.46} & \textbf{63.07} & 34.24 & \textbf{67.31} & 58.00 & \underline{1.114} & \underline{2.5} \\
& CompassMem & 22.40 & 29.20 & 22.40 & \underline{58.00} & 24.60 & 30.60 & 54.30 & 62.80 & 56.00 & \underline{68.00} & 65.10 & \underline{61.00} & 1.032 & 4.2 \\
& FTA-Mem    & \underline{37.31} & 29.99 & \textbf{31.11} & \textbf{72.80} & 24.87 & \textbf{38.52} & \textbf{61.79} & 60.61 & \underline{62.62} & \textbf{81.75} & 63.07 & \textbf{65.64} & \textbf{1.121} & \textbf{2.4} \\

\midrule
\multirowcell{6}{GPT-4o-mini}
& MemoryBank & 25.85 & \underline{32.35} & 25.01 & 57.16 & \underline{26.62} & 32.88 & 54.70 & \underline{65.07} & 58.23 & 68.83 & \textbf{66.43} & \underline{62.52} & \underline{1.112} & 2.8 \\
& MemGPT     & \underline{29.63} & \textbf{34.99} & \underline{30.21} & 45.46 & \textbf{27.60} & 33.26 & \underline{55.94} & \textbf{66.77} & \textbf{62.82} & 59.84 & \underline{66.38} & 62.34 & 1.080 & \underline{2.4} \\
& MemoryOS   & 24.80 & 28.10 & 20.90 & 65.00 & 24.70 & 32.06 & 54.70 & 62.60 & 55.94 & 68.10 & 65.45 & 61.26 & 1.020 & 4.4 \\
& A-Mem      & 26.33 & 30.77 & 22.09 & \underline{66.95} & 26.40 & \underline{33.86} & 54.86 & 63.15 & 53.93 & \underline{76.36} & 64.82 & 62.40 & 1.099 & 3.1 \\
& CompassMem & 24.60 & 30.40 & 23.20 & 62.50 & 25.10 & 31.80 & 55.00 & 63.60 & 57.00 & 72.00 & 65.20 & 62.00 & 1.054 & 4.0 \\
& FTA-Mem    & \textbf{40.93} & 29.50 & \textbf{31.25} & \textbf{68.01} & 25.68 & \textbf{38.53} & \textbf{65.22} & 61.83 & \underline{62.80} & \textbf{78.08} & 64.04 & \textbf{66.19} & \textbf{1.122} & \textbf{2.3} \\

\bottomrule
\end{tabular}}
\caption{Performance on ES-MemEval. Avg.R is computed over the twelve automatic metrics, i.e., F1 and BERTScore on IE/TR/CD/Abs/UM/All; Judge is not included in Avg.R. Best results are in bold, and second-best results are underlined.}
\label{tab:es_memeval_main}
\end{table*}

% \midrule

% \multirow{5}{*}{Qwen3-4B}
% & MemoryBank & 10.60 & 13.14 & 24.15 & 20.28 & \textbf{25.24} & \textbf{20.92} & 14.23 & 10.53 & 3.2 & 3.2 \\
% & MemGPT     & 3.11 & 2.28 & 2.52 & 1.56 & 0.93 & 0.61 & 8.14 & 6.14 & 5.0 & 5.0 \\
% & MemoryOS   & 17.79 & \textbf{22.29} & 24.67 & 20.90 & 3.01 & 2.16 & \textbf{17.30} & \textbf{14.61} & 2.8 & 2.0 \\
% & A-Mem      & \textbf{20.44} & 16.81 & 25.47 & 20.83 & 23.20 & 18.29 & 16.25 & 12.68 & 2.2 & 3.0 \\
% & FTA-Mem    & 19.79 & 21.45 & \textbf{28.18} & \textbf{23.64} & 24.69 & 20.49 & 16.48 & 13.02 & \textbf{1.8} & \textbf{1.8} \\

\begin{table*}[t]
\centering
{\small
\setlength{\tabcolsep}{4pt}
\renewcommand{\arraystretch}{1.08}
\begin{tabular}{c l|cc|cc|cc|cc|cc}
\toprule
\multirow{2}{*}{\textbf{Backbone}}
& \multirow{2}{*}{\textbf{Method}}
& \multicolumn{2}{c|}{\textbf{Single Hop}}
& \multicolumn{2}{c|}{\textbf{Multi Hop}}
& \multicolumn{2}{c|}{\textbf{Temporal}}
& \multicolumn{2}{c|}{\textbf{Open Domain}}
& \multicolumn{2}{c}{\textbf{Average} $\uparrow$} \\
& & F1 $\uparrow$ & BLEU-1 $\uparrow$
& F1 $\uparrow$ & BLEU-1 $\uparrow$
& F1 $\uparrow$ & BLEU-1 $\uparrow$
& F1 $\uparrow$ & BLEU-1 $\uparrow$
& F1 & BLEU-1 \\

\midrule
\multirow{6}{*}{Qwen3-8B}
& MemoryBank & 14.32 & 16.07 & 28.30 & 23.78 & 22.25 & 18.69 & 15.32 & 12.33 & 20.05 & 17.72 \\
& MemGPT     & 21.66 & 22.43 & 19.14 & 15.28 & 3.71 & 2.81 & 11.86 & 9.28 & 14.09 & 12.45 \\
& MemoryOS   & 18.65 & 22.67 & 26.07 & 22.31 & 3.62 & 2.52 & 14.21 & 11.44 & 15.64 & 14.74 \\
& A-Mem      & 23.92 & 22.69 & 29.42 & 24.85 & 26.81 & 21.85 & 15.68 & 11.90 & 23.96 & 20.32 \\
& CompassMem & \textbf{31.73} & \textbf{29.12} & \textbf{42.05} & \textbf{35.86} & \underline{34.18} & \underline{28.06} & \underline{18.64} & \underline{14.67} & \underline{31.65} & \underline{26.93} \\
& FTA-Mem    & \underline{31.16} & \underline{26.62} & \underline{41.84} & \underline{35.75} & \textbf{35.93} & \textbf{29.90} & \textbf{20.86} & \textbf{16.62} & \textbf{32.45} & \textbf{27.22} \\

\midrule
\multirow{6}{*}{Qwen3-14B}
& MemoryBank & 11.96 & 12.65 & 25.37 & 21.03 & 20.06 & 16.06 & 12.17 & 9.41 & 17.39 & 14.79 \\
& MemGPT     & 21.35 & 17.42 & 22.32 & 18.64 & 6.34 & 5.42 & 11.52 & 8.64 & 15.38 & 12.53 \\
& MemoryOS   & 17.58 & 20.34 & 26.34 & 22.13 & 5.11 & 3.40 & 13.07 & 10.51 & 15.53 & 14.09 \\
& A-Mem      & 23.68 & 22.88 & 28.31 & 24.03 & 21.06 & 16.30 & \textbf{16.16} & \textbf{12.62} & 22.30 & 18.96 \\
& CompassMem & \textbf{28.40} & \textbf{27.50} & \textbf{39.50} & \textbf{33.60} & \underline{23.40} & \underline{18.40} & 12.40 & 8.70 & \underline{25.93} & \underline{22.05} \\
& FTA-Mem    & \underline{27.84} & \underline{26.22} & \underline{39.32} & \underline{33.52} & \textbf{25.19} & \textbf{20.23} & \underline{14.58} & \underline{10.66} & \textbf{26.73} & \textbf{22.66} \\

\midrule
\multirow{6}{*}{GPT-4o-mini}
& MemoryBank & 10.81 & 9.29 & 22.35 & 17.66 & 19.64 & 14.39 & 11.87 & 9.16 & 16.17 & 12.62 \\
& MemGPT     & 22.65 & 18.72 & 24.45 & 19.44 & 9.15 & 7.44 & 11.04 & 8.34 & 16.82 & 13.48 \\
& MemoryOS   & 14.61 & 16.06 & 24.95 & 20.69 & 4.44 & 3.36 & \underline{12.37} & \underline{9.93} & 14.09 & 12.51 \\
& A-Mem      & 17.95 & 14.50 & 24.98 & 20.15 & 21.98 & 16.69 & \textbf{14.56} & \textbf{11.42} & 19.87 & 15.69 \\
& CompassMem & \textbf{23.60} & \textbf{23.70} & \textbf{37.00} & \textbf{30.50} & \underline{29.20} & \underline{23.70} & 9.20 & 6.80 & \underline{24.75} & \underline{21.18} \\
& FTA-Mem    & \underline{23.01} & \underline{21.16} & \underline{36.74} & \underline{30.38} & \textbf{30.95} & \textbf{25.52} & 11.41 & 8.75 & \textbf{25.53} & \textbf{21.45} \\

\bottomrule
\end{tabular}}
\caption{Results on LoCoMo category-level long-term memory question answering. Best results are in bold, and second-best results are underlined.}
\label{tab:locomo_category_main}
\end{table*}

% \begin{table}[t]
% \centering
% \small
% \setlength{\tabcolsep}{4pt}
% \renewcommand{\arraystretch}{1.08}
% \begin{tabular}{lrrr}
% \toprule
% \textbf{Method}
% & \textbf{Token/Q}
% & \textbf{F1 $\uparrow$}
% & \textbf{B/B1 $\uparrow$} \\
% \midrule

% \multicolumn{4}{l}{\textit{\textbf{ES-MemEval}}} \\
% MemoryBank      & 982 & 32.88 & 62.52 \\
% MemGPT          & 3,003 & 33.26 & 62.34 \\
% MemoryOS        & 3,658 & 32.06 & 61.26 \\
% A-Mem         & 3,103 & 33.86 & 62.40 \\
% CompassMem      & 1,268 & 30.20 & 60.67 \\
% FTA-Mem         & \textbf{2,084} & \textbf{38.53} & \textbf{66.19} \\

% \midrule
% \multicolumn{4}{l}{\textit{\textbf{LoCoMo}}} \\
% MemoryBank      & 612 & 34.69 & 31.50 \\
% MemGPT          & 4,084 & 24.71 & 23.96 \\
% MemoryOS        & 3,874 & 35.31 & 33.41 \\
% A-Mem          & 2,976 & 37.88 & 34.34 \\
% CompassMem      & 1,467 & \textbf{31.65 } & \textbf{31.68} \\
% FTA-Mem         & \textbf{1,518} & \textbf{45.63} & \textbf{41.67} \\

% \bottomrule
% \end{tabular}
% \caption{Answer-stage cost and average performance. Token/Q denotes response-generation tokens per question. }
% \label{tab:answer_cost}
% \end{table}

\begin{table}[t]
\centering
{\small
\setlength{\tabcolsep}{4pt}
\renewcommand{\arraystretch}{1.08}
\begin{tabular}{lrrrr}
\toprule
\textbf{Gain} & \textbf{Fact} & \textbf{Ent.} & \textbf{Time} & \textbf{Impl.} \\
\midrule
\multicolumn{5}{l}{\textit{\textbf{ES-MemEval}}} \\
FTA $-$ CompassMem & -0.39 & -0.12 & -0.14 & 0.41 \\
FTA $-$ Avg. & -0.32 & -0.07 & -0.08 & 0.44 \\
\midrule
\multicolumn{5}{l}{\textit{\textbf{LoCoMo}}} \\
FTA $-$ CompassMem & 0.00 & 0.47 & -0.44 & 0.21 \\
FTA $-$ Avg.  & 0.26 & 0.40 & -0.12 & -0.10 \\
\bottomrule
\end{tabular}}
\caption{Diagnostic Pearson correlations between dialogue-level density fields and FTA-Mem's relative F1 gain. Fact, entity, and time densities are mean annotated anchors per turn. Avg. denotes the mean over MemoryBank, MemGPT, MemoryOS, A-Mem, and CompassMem.}
\label{tab:density_gain_corr}
\end{table}

\begin{table}[t]
\centering
{\small
\setlength{\tabcolsep}{4pt}
\renewcommand{\arraystretch}{1.08}
\begin{tabular}{llrr}
\toprule
\textbf{Method}
& \textbf{Retrieval Unit}
& \textbf{Units}
& \textbf{Token/Q} \\
\midrule

\multicolumn{4}{l}{\textit{\textbf{ES-MemEval}}} \\
MemoryBank & Summary/profile & 838 & 982 \\
MemGPT     & Archival passage & 401 & 3,003 \\
MemoryOS   & Turn-level state & 4,321 & 3,658 \\
A-Mem      & Agentic note & 401 & 3,103 \\
CompassMem & Event graph node & 2,642 & 1,268 \\
FTA-Mem    & Situation-level FTA & 6,564 & 2,084 \\

\midrule
\multicolumn{4}{l}{\textit{\textbf{LoCoMo}}} \\
MemoryBank & Summary/profile & 564 & 615 \\
MemGPT     & Archival passage & 272 & 4,915 \\
MemoryOS   & Turn-level state & 2,807 & 534 \\
A-Mem      & Agentic note & 272 & 2,978 \\
CompassMem & Event graph node & 2,190 & 1,467 \\
FTA-Mem    & Situation-level FTA & 4,299 & 1,713 \\

\bottomrule
\end{tabular}}
\caption{Answer-stage cost comparison. Token/Q denotes response-generation tokens per question.}
\label{tab:answer_cost}
\end{table}

\section{Experiments}

\subsection{Experimental Settings}

\subsubsection{Datasets.}
We evaluate FTA-Mem on two long-term dialogue memory benchmarks. ES-MemEval focuses on long-term emotional-support interactions and contains 18 user cases with 1,427 questions. LoCoMo is a general long conversational memory benchmark with 10 conversation samples and 1,986 questions, covering single-hop, multi-hop, temporal, and open-domain reasoning.

To characterize their difference, we conduct a turn-level information-density analysis in Table~\ref{tab:density_analysis}. Specifically, we use GPT-5.5 to annotate each dialogue turn with our density prompt, counting factual, entity, temporal, and support-process anchors, rating implicitness on a 1--3 scale, and assigning a binary low-information label. We then report dataset-level averages. LoCoMo contains more explicit factual, entity, temporal, and total anchors per turn, indicating denser retrievable evidence. In contrast, ES-MemEval contains more support-process signals and higher implicitness, suggesting that its useful evidence is more often embedded in supportive interactions and depends more on dialogue context.

\subsubsection{Metrics.}
For ES-MemEval, we report F1 and BERTScore over five capabilities~\citep{zhang2020bertscoreevaluatingtextgeneration}: information extraction (IE), temporal reasoning (TR), conflict detection (CD), abstention (Abs), and user modeling (UM). We also report an average LLM-as-a-judge score on a 0--2 scale, where GPT-5.5 is used as the judge model. For LoCoMo, we report F1 and BLEU-1 for Single Hop, Multi Hop, Temporal, and Open Domain questions. We additionally report average rank to summarize robustness across question types.

\subsubsection{Baselines.}
We compare FTA-Mem with representative long-term memory systems for LLM agents. MemoryBank stores dialogue history through summaries and user-related memories. MemGPT, implemented with Letta, represents an external-memory agent baseline with archival memory retrieval. MemoryOS maintains hierarchical memory states for long-term personalization. A-Mem is an agentic memory baseline with memory construction, linking, and update. CompassMem constructs event-centric memory graphs and hierarchical topic structures. We also include question-only, fact-only, and component-ablated variants in the ablation studies.
\\
\textbf{Implementation Details.}
FTA-Mem uses FTA units with \(W=20\) for BWS. Retrieval uses all-MiniLM-L6-v2 embeddings with a threshold of \(\tau=0.35\)~\citep{wang2020minilm}, and we set \(\lambda=0.8\) in Eq.~(\ref{eq:retrieval_score}). The retrieval budget is \(K=10\) on ES-MemEval and \(K=25\) on LoCoMo. Query rewriting is enabled, and one linked neighbor is added per retrieved unit. Qwen models are served with vLLM on two NVIDIA A800 GPUs~\citep{yang2025qwen3technicalreport, kwon2023efficient}; GPT-4o-mini and GPT-5.5 judge evaluation use the API. All answer generation uses temperature 0.

\subsection{Main Results}

\subsubsection{Overall Performance}

Table~\ref{tab:es_memeval_main} reports the results on ES-MemEval. FTA-Mem achieves the strongest average performance across backbone settings. With Qwen3-8B, it obtains 38.71 F1, 66.68 BERTScore, and the best average rank; with GPT-4o-mini, it also achieves the best overall F1 and BERTScore. The gains are most visible in information extraction and conflict detection, indicating that FTA units better preserve answerable evidence together with temporal and affective context.

Existing memory baselines show less stable behavior across capabilities. MemoryOS and CompassMem perform strongly on abstention-related cases, while MemGPT and A-Mem are competitive on some individual categories. However, they do not consistently perform well across IE, temporal reasoning, conflict detection, and user modeling. This suggests that low-density emotional-support memory requires more than generic summaries, profiles, event localization, or external-memory retrieval; the retrievable unit itself should encode factual evidence, temporal validity, and affective context.

Table~\ref{tab:locomo_category_main} shows results on LoCoMo. FTA-Mem achieves the best average F1 and BLEU-1 across all backbone settings, suggesting that situation-level FTA units remain competitive and effective in denser long-term dialogue. On Qwen3-8B, CompassMem is slightly stronger on single-hop and multi-hop questions, while FTA-Mem performs best on temporal and open-domain questions and obtains the best overall average. This pattern is consistent with Table~\ref{tab:density_analysis}: LoCoMo contains denser factual and entity anchors, so event-centric or conventional retrieval remains competitive, while FTA-Mem brings clearer gains when temporal state and distributed evidence matter.

\subsubsection{Granularity Analysis}

Table~\ref{tab:granularity_cost} compares construction granularities. On ES-MemEval, session-level construction (\(W=\) Session) is too coarse and loses useful evidence, while turn-pair construction (\(W=2\)) creates more units and calls but still underperforms situation-level FTA construction. On LoCoMo, turn-pair construction is slightly stronger, but requires substantially higher construction cost. These results indicate that very fine-grained construction can help in denser factual settings, but it is less efficient and more redundant. Situation-level construction provides a better overall trade-off. A controlled comparison with fixed-window segmentation is provided in the supplementary material.

\subsubsection{Answer-Stage Cost}

Table~\ref{tab:answer_cost} reports answer-stage cost. FTA-Mem constructs fewer units than turn-level or turn-pair memory construction, but more units than summary-based or coarse event-localization methods. This reflects its situation-level design: it preserves sparse evidence at a finer granularity than summaries, while avoiding the redundancy of indexing every turn. Despite using richer memory units, FTA-Mem keeps Token/Q within a competitive range despite using richer structured units.

\subsubsection{Density-Gain Correlation}

Table~\ref{tab:density_gain_corr} analyzes when FTA-Mem obtains larger relative gains. On ES-MemEval, the gain over CompassMem is higher when factual density is lower (\(r=-0.39\)) and implicitness is higher (\(r=0.41\)); the average-baseline comparison shows the same trend. Entity and time densities are less correlated with the gain.
This pattern is not consistent on LoCoMo, suggesting that the density-related advantage is more specific to ES-MemEval. Overall, the results provide diagnostic evidence that FTA-Mem is especially useful when explicit factual evidence is sparse and user meaning is implicit.

% \begin{table*}[t]
% \centering
% \small
% \setlength{\tabcolsep}{4pt}
% \renewcommand{\arraystretch}{1.08}
% \begin{tabular}{llrrrrrr}
% \toprule
% \textbf{Benchmark} & \textbf{Construction Unit}
% & \textbf{F1 $\uparrow$}
% & \textbf{BERT/BLEU $\uparrow$}
% & \textbf{Units}
% & \textbf{Unit Calls}
% & \textbf{Unit Tokens}
% & \textbf{Tok./Unit} \\
% \midrule

% \multirow{3}{*}{ES-MemEval}
% & Session-level        & 31.76 & 61.78 & 1,692 & 401 & 1.58M & 935 \\
% & Turn-pair            & 37.06 & 65.43 & 8,556 & 4,883 & 6.40M & 748 \\
% & FTA-Mem& \textbf{38.71} & \textbf{66.68} & 6,564 & 3,245 & 4.99M & 760 \\

% \midrule
% \multirow{3}{*}{LoCoMo}
% & Session-level        & 29.34 & 25.13 & 1,008 & 272   & 1.09M & 1,081 \\
% & Turn-pair            & \textbf{38.28} & \textbf{32.51} & 6,071 & 5,882 & 7.04M & 1,159 \\
% & FTA-Mem& 37.35 & 31.67 & 4,299 & 2,061 & 3.39M & 788 \\

% \bottomrule
% \end{tabular}
% \caption{Granularity and memory-unit construction cost. Units denotes the total number of constructed memory units. Unit Calls and Unit Tokens count only LLM calls and tokens used for memory-unit construction. Tok./Unit is computed as Unit Tokens divided by Units.}
% \label{tab:granularity_cost}
% \end{table*}

\begin{table}[t]
\centering
{\small
\setlength{\tabcolsep}{3pt}
\renewcommand{\arraystretch}{1.08}
\begin{tabular}{lrrrrr}
\toprule
\textbf{Unit}
& \textbf{F1 $\uparrow$}
& \textbf{B/B1 $\uparrow$}
& \textbf{Units}
& \textbf{Calls}
& \textbf{Tokens} \\
\midrule

\multicolumn{6}{l}{\textit{ES-MemEval}} \\
Session-level & 31.76 & 61.78 & 1,692 & 401 & 1.58M \\
Turn-pair     & 37.06 & 65.43 & 8,556 & 4,883 & 6.40M \\
FTA-Mem       & \textbf{38.71} & \textbf{66.68} & 6,564 & 3,245 & 4.99M \\

\midrule
\multicolumn{6}{l}{\textit{LoCoMo}} \\
Session-level & 29.34 & 25.13 & 1,008 & 272   & 1.09M \\
Turn-pair     & \textbf{38.28} & \textbf{32.51} & 6,071 & 5,882 & 7.04M \\
FTA-Mem       & 37.35 & 31.67 & 4,299 & 2,061 & 3.39M \\

\bottomrule
\end{tabular}}
\caption{Granularity and memory-unit construction cost. Units is the total number of constructed memory units. Calls and Tokens count only LLM calls and tokens used for memory-unit construction.}
\label{tab:granularity_cost}
\end{table}

\begin{figure}[t]
\centering
\includegraphics[width=\columnwidth]{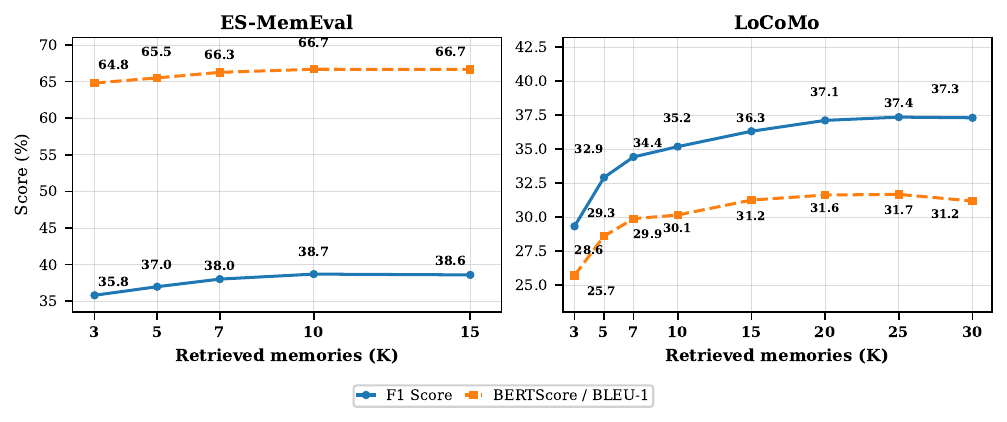} 
\caption{Effect of retrieval budget K on ES-MemEval and LoCoMo.}
\label{fig:k_sensitivity}
\end{figure}

\begin{figure}[t]
\centering
\includegraphics[width=\columnwidth]{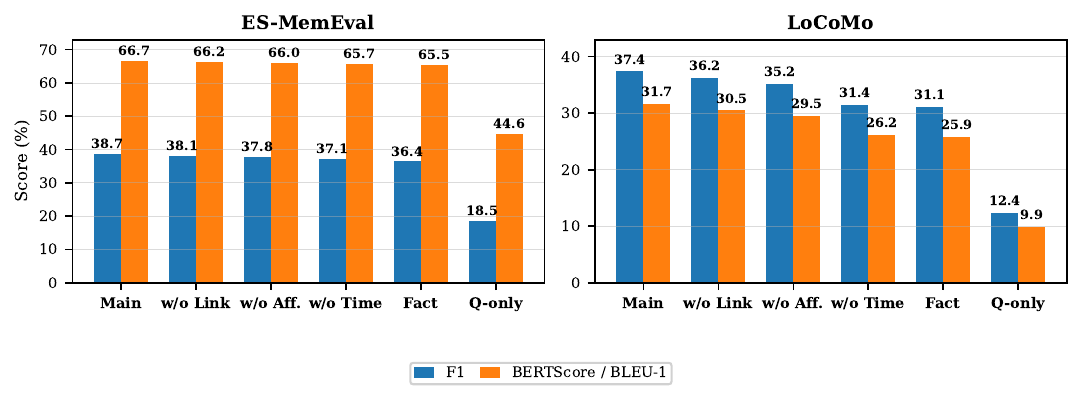} 
\caption{Ablation results on ES-MemEval and LoCoMo.}
\label{fig:ablation_bars}
\end{figure}

\subsection{Hyperparameter Analysis}

Figure~\ref{fig:k_sensitivity} analyzes the effect of the retrieval budget \(K\). On ES-MemEval, performance increases from \(K=3\) to \(K=10\), reaching 38.71 F1 and 66.68 BERTScore, and then saturates. This suggests that ES-MemEval needs enough structured evidence, but overly large retrieval sets bring limited benefit.

On LoCoMo, pooled over Cat1--Cat4, performance improves more gradually and peaks around \(K=25\). Since LoCoMo contains denser entity and factual cues, a larger retrieval budget can provide useful additional evidence. This contrast supports our density analysis: low-density emotional-support dialogue benefits more from selective retrieval and structured context synthesis, while denser factual dialogue can tolerate larger retrieval sets.

\subsection{Ablation Study}

Figure~\ref{fig:ablation_bars} presents the ablation results. Removing temporal information causes the largest degradation, especially on LoCoMo, showing the importance of temporal grounding for evolving situations. Removing links causes a smaller drop, suggesting that relation maintenance provides auxiliary longitudinal cues. Removing affect information also degrades performance. Since the affect anchor includes emotion, intention, and relation cues, this effect can extend beyond emotional-support questions to person-relation and plan-related reasoning.

Fact-Only memory also underperforms the full FTA unit on both benchmarks, indicating that factual content alone is insufficient for long-term dialogue memory. The Question-only baseline performs much worse, confirming that the gains come from retrieved memory rather than backbone knowledge alone. Overall, FTA-Mem benefits from combining factual evidence, temporal grounding, affective context, and lightweight links. Further analyses are provided in the supplementary material.

\section{Conclusion}

We propose FTA-Mem, a structured memory framework for low-density long-term dialogue. FTA-Mem constructs situation-level Fact-Time-Affect units through boundary-preserving segmentation, preserving factual evidence together with temporal and affective context. It further organizes these units with temporal links and structured context synthesis to support grounded and personalized answer generation. Experiments on ES-MemEval and LoCoMo show that FTA-Mem improves long-term memory question answering across different dialogue settings and provides a favorable trade-off among evidence preservation, temporal consistency, and construction efficiency.

\bibliography{aaai2027}


\section{Supplementary Material}

\subsection{Reproducibility Details}

Answer generation uses temperature 0, while memory construction uses
temperature 0.7. In FTA-Mem, BWS, FTA-unit extraction, relation
classification, query rewriting, and answer generation are implemented as
separate LLM calls with fixed prompts. Local Qwen backbones are served with
vLLM on two NVIDIA A800 GPUs; GPT-4o-mini inference and GPT-5.5-based judge
evaluation are conducted through APIs. Retrieval uses all-MiniLM-L6-v2 embeddings with cosine
similarity. For all compared methods, we use the same answer backbone and
generation settings while preserving each baseline's native memory format
and context construction procedure. Unless otherwise specified, all
additional analyses and ablation experiments use Qwen3-8B as the backbone.

For retrieval, we use the same top-$K$ protocol across methods
when applicable. Guided by the default retrieval settings reported
in the original benchmark studies, the main comparative experiments
(Tables~2--3) use $K=7$ for ES-MemEval and $K=10$ for
LoCoMo. Following the top-$K$ sensitivity analysis, we use
$K=10$ for ES-MemEval and $K=25$ for LoCoMo in the subsequent
ablation, granularity, and diagnostic experiments
(Tables~4--6). Token/Q denotes the resulting answer-stage tokens per
question under each method's native memory representation and
context-construction procedure, rather than a manually matched token budget.

\subsection{Baseline Selection}
ES-Mem is not included as a baseline because it had not been formally accepted and no official implementation was publicly available at the time of submission. SGMem is likewise excluded because no official implementation was publicly available. Moreover, both methods are closely related to fine-grained turn- or sentence-level memory construction, which is already examined in our granularity comparison.

\subsection{Boundary-preserving Window Segmentation}

Algorithm~\ref{alg:bws} gives the BWS procedure. The detector returns segment boundaries using the global numeric turn indices in the session. For each window, BWS finalizes completed segments before the last detected segment and keeps the last segment provisional. The next window starts from the beginning of this provisional segment, allowing boundary-tail content to be reprocessed with following context. If the detector returns a single full-window segment, BWS enforces a minimum forward step to avoid non-termination. When the window reaches the end of the session, all remaining segments are finalized.

\begin{algorithm}
\caption{Boundary-preserving Window Segmentation}
\label{alg:bws}
\begin{algorithmic}[1]
\REQUIRE Session $S$, window size $W$
\STATE $s \leftarrow 0$
\STATE $\mathcal{F} \leftarrow \emptyset$
\WHILE{$s < |S|$}
    \STATE $E \leftarrow \min(s+W-1, |S|-1)$
    \STATE $\{(b_j,e_j,r_j)\}_{j=1}^{K}
    \leftarrow g_\phi(S[s:E])$
    \IF{$E = |S|-1$}
        \STATE add all detected segments to $\mathcal{F}$
        \STATE \textbf{break}
    \ELSE
        \STATE add segments $0,\ldots,K-2$ to $\mathcal{F}$ if $K>1$
        \STATE retain the last segment as provisional
        \STATE $s \leftarrow \max(b_{K-1}, s+1)$
    \ENDIF
\ENDWHILE
\RETURN $\mathcal{F}$
\end{algorithmic}
\end{algorithm}

\subsection{Additional Comparative Experiments}

To directly examine the effect of boundary-preserving segmentation, we compare
BWS with a fixed-window variant on ES-MemEval and LoCoMo. The fixed-window
setting uses the same window size, memory-unit schema, retriever, and
answer-generation pipeline, but advances windows with a fixed stride and does
not carry the last boundary-uncertain segment into the next window. Therefore,
the comparison isolates the effect of preserving segment boundaries during
memory construction.

\begin{table}[t]
\centering
\small
\setlength{\tabcolsep}{6pt}
\renewcommand{\arraystretch}{1.08}
\begin{tabular}{llccc}
\toprule
\textbf{Dataset} & \textbf{Setting} & \textbf{F1 $\uparrow$} & \textbf{B/B1 $\uparrow$} & \textbf{Units} \\
\midrule
ES-MemEval & BWS & \textbf{38.71} & \textbf{66.68} & \textbf{6,564} \\
ES-MemEval & Fixed Window & 37.94 & 65.34 & 6,678 \\
\midrule
LoCoMo & BWS & \textbf{37.35} & \textbf{31.67} & \textbf{4,299} \\
LoCoMo & Fixed Window & 37.04 & 30.84 & 4,332 \\
\bottomrule
\end{tabular}
\caption{Comparison between boundary-preserving window segmentation and fixed-window segmentation. B/B1 denotes BERTScore on ES-MemEval and BLEU-1 on LoCoMo. Scores are reported in percentages.}
\label{tab:bws_fixed_window}
\end{table}

The fixed-window variant constructs slightly more memory units but obtains
lower F1 and B/B1 on both benchmarks. This suggests that simply increasing the
number of segments is not sufficient; preserving uncertain boundary-tail
content helps avoid incomplete memory units and improves downstream question
answering.

\subsection{Additional Ablation Studies}

Table~\ref{tab:es_additional_ablation} reports additional ablations on
ES-MemEval. Removing query rewriting leads to a moderate drop, suggesting
that evidence-oriented reformulation helps align user questions with
structured FTA units. Plain embedding retrieval further reduces performance,
showing that structured cue matching over time, situation type, participants,
and affective context contributes beyond semantic similarity alone.

Replacing the structured memory packet with flat context also degrades
performance. This indicates that FTA-Mem benefits not only from retrieving
relevant units, but also from organizing them into primary evidence, linked
neighbors, source spans, and auxiliary context. In contrast, removing
auxiliary memory causes little change, suggesting that episodic FTA units
remain the primary evidence source, while user semantic memory and
support-experience memory mainly provide background personalization.

\begin{table}[t]
\centering
\small
\setlength{\tabcolsep}{5pt}
\renewcommand{\arraystretch}{1.08}
\begin{tabular}{lcc}
\toprule
\textbf{Setting} & \textbf{F1 $\uparrow$} & \textbf{BERTScore $\uparrow$} \\
\midrule
FTA-Mem & \textbf{38.71} & \textbf{66.68} \\
w/o Query Rewrite & 37.44 & 65.90 \\
Plain Embedding & 35.99 & 64.88 \\
Flat Context & 37.86 & 65.24 \\
w/o Auxiliary Memory & 38.65 & 66.73 \\
\bottomrule
\end{tabular}
\caption{Additional ablation results on ES-MemEval. Scores are reported in percentages.}
\label{tab:es_additional_ablation}
\end{table}

\subsection{Density Annotation and Statistical Analysis}

Each dialogue turn is annotated with factual, entity, temporal, and
support-process anchor counts. Anchor fields are non-negative integer
counts; implicitness is rated on a 1--3 scale, and low-information status
is binary. We aggregate turn-level annotations into dialogue-level
statistics by averaging over turns.

We further report the same density-gain correlation setting used in the main
paper. For each ES-MemEval case or LoCoMo conversation, we compute the
relative F1 gain of FTA-Mem and correlate it with dialogue-level density
fields. Table~\ref{tab:density_gain_robustness} reports Pearson \(r\) for
the two main diagnostic fields, together with auxiliary robustness statistics.

\begin{table*}[t]
\centering
\small
\setlength{\tabcolsep}{4pt}
\renewcommand{\arraystretch}{1.05}
\begin{tabular}{lllrrrrr}
\toprule
\textbf{Dataset} & \textbf{Gain} & \textbf{Field}
& \textbf{Pearson \(r\)}
& \textbf{Spearman \(\rho\)}
& \textbf{95\% CI}
& \textbf{Perm. \(p\)}
& \textbf{LOO Range} \\
\midrule
ES-MemEval & FTA--CompassMem & Fact
& -0.39 & -0.48 & [-0.70, -0.04] & 0.110 & [-0.49, -0.32] \\
ES-MemEval & FTA--CompassMem & Implicit.
& 0.43 & 0.41 & [0.00, 0.73] & 0.077 & [0.31, 0.53] \\
ES-MemEval & FTA--Avg. Baseline & Fact
& -0.32 & -0.36 & [-0.74, 0.13] & 0.154 & [-0.54, -0.20] \\
ES-MemEval & FTA--Avg. Baseline & Implicit.
& 0.44 & 0.34 & [-0.15, 0.78] & 0.081 & [0.23, 0.55] \\
\midrule
LoCoMo & FTA--CompassMem & Fact
& 0.00 & -0.20 & [-0.77, 0.60] & 0.481 & [-0.53, 0.07] \\
LoCoMo & FTA--CompassMem & Implicit.
& 0.21 & 0.31 & [-0.16, 0.96] & 0.306 & [0.26, 0.63] \\
LoCoMo & FTA--Avg. Baseline & Fact
& 0.26 & 0.21 & [-0.66, 0.74] & 0.694 & [-0.08, 0.34] \\
LoCoMo & FTA--Avg. Baseline & Implicit.
& -0.10 & -0.16 & [-0.64, 0.72] & 0.998 & [-0.13, 0.21] \\
\bottomrule
\end{tabular}
\caption{Additional density-gain correlations under the same setting as the
main-paper diagnostic table. Pearson \(r\) follows the main-paper aggregation.
CI denotes case-level bootstrap 95\% confidence interval; Perm. \(p\) is
computed by permutation testing; and LOO Range reports leave-one-case-out
Pearson ranges.}
\label{tab:density_gain_robustness}
\end{table*}

On ES-MemEval, the diagnostic trend suggests that FTA-Mem gains more when
factual density is lower and implicitness is higher. The result is best
viewed as diagnostic evidence rather than a causal claim. The pattern is
weaker on LoCoMo, suggesting that the density-related advantage is more
evident in the low-density emotional-support benchmark.

\begin{figure}[t]
\centering
\includegraphics[width=0.48\columnwidth]{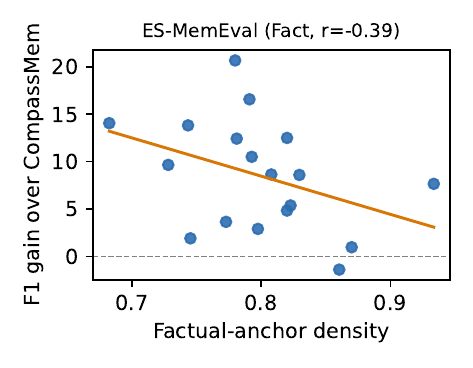}
\includegraphics[width=0.48\columnwidth]{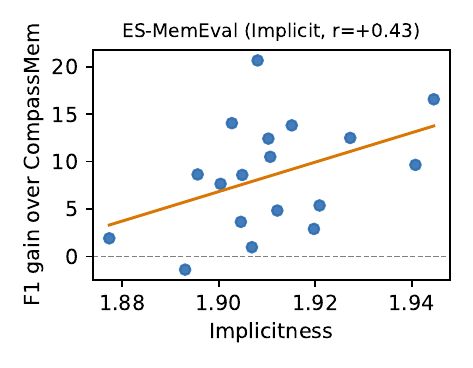}
\caption{Dialogue-level density-gain trends on ES-MemEval.}
\label{fig:density_gain_scatter}
\end{figure}

\subsection{Paired Bootstrap Uncertainty Check}

We estimate the uncertainty of the main FTA-Mem--CompassMem comparison with
paired cluster bootstrap. ES-MemEval cases and LoCoMo conversations are
resampled with replacement, and the F1 difference is averaged over the
sampled clusters. This setting reflects the multi-question structure of
each dialogue case while keeping the constructed memories fixed.

As shown in
Table~\ref{tab:paired_bootstrap}, the ES-MemEval gain remains clearly
positive, while the LoCoMo interval is much closer to zero. This suggests
that the advantage over CompassMem is more stable on ES-MemEval, whereas
the LoCoMo difference should be interpreted cautiously.

\begin{table}[t]
\centering
\small
\setlength{\tabcolsep}{5pt}
\renewcommand{\arraystretch}{1.08}
\begin{tabular}{lrrr}
\toprule
\textbf{Dataset} & \textbf{Clusters} & \textbf{Mean Diff.} & \textbf{95\% CI} \\
\midrule
ES-MemEval & 18 & 8.43 & [5.81, 11.04] \\
LoCoMo & 10 & 0.55 & [-0.01, 3.18] \\
\bottomrule
\end{tabular}
\caption{Paired cluster bootstrap uncertainty check for FTA-Mem minus
CompassMem. Values are token-level F1 differences in percentage points.
Clusters correspond to ES-MemEval cases or LoCoMo conversations. LoCoMo
excludes Cat5 no-information questions.}
\label{tab:paired_bootstrap}
\end{table}

\subsection{Memory Quality Audit}

To examine whether the intermediate memory representation is grounded before
retrieval and answer generation, we conduct a small-scale manual audit of
FTA-Mem outputs. We randomly sample 100 generated FTA units from the
Qwen3-8B runs, balanced across ES-MemEval and LoCoMo. For each unit, two
annotators independently evaluate factual grounding, temporal accuracy, and
affect/context accuracy by comparing the memory fields with the cited
evidence span, and we report the mean of their scores. We additionally
sample 50 linked pairs, balanced across the two benchmarks, and evaluate
whether the linked units are meaningfully related for memory maintenance and
retrieval.

Each criterion is scored on a 0--2 scale: 2 denotes fully correct and
grounded, 1 denotes partially correct but usable, and 0 denotes incorrect or
unsupported. Table~\ref{tab:memory_quality_audit} reports the average score
for each criterion.

\begin{table}[t]
\centering
\small
\setlength{\tabcolsep}{4pt}
\renewcommand{\arraystretch}{1.08}
\begin{tabular}{lcccc}
\toprule
\textbf{Benchmark} & \textbf{Fact} & \textbf{Time} & \textbf{Affect/Ctx.} & \textbf{Rel.} \\
\midrule
ES-MemEval & 2.00 & 2.00 & 1.48 & 1.36 \\
LoCoMo & 2.00 & 2.00 & 1.38 & 1.25 \\
\bottomrule
\end{tabular}
\caption{Manual audit of intermediate FTA memory quality. Scores are reported on a 0--2 scale. Rel. denotes relation relatedness for linked memory pairs.}
\label{tab:memory_quality_audit}
\end{table}

The audit suggests that FTA units are strongly grounded in factual and
temporal evidence on both benchmarks. Affect/context information is usually
usable but less often fully explicit, reflecting that emotional state and
intentions are sometimes inferred from local context. Relation relatedness
is also useful but noisier than the unit-level anchors, which supports our
treatment of links as lightweight maintenance cues rather than a fully
precise symbolic graph.

\subsection{Case Study}
\subsubsection{Unit Case}

Figure~\ref{fig:unit_case} shows a representative FTA unit. The example illustrates how a situation-level unit stores factual, temporal, affective, evidence, and link-maintenance fields.

\begin{figure}[t]
\centering
\small
\fbox{
\begin{minipage}{0.93\columnwidth}
\textbf{Fact-Time-Affect Memory Unit}

\vspace{2pt}
\textbf{memory\_id}: emem\_000021

\textbf{summary}: Seeker mentions having friends but feels they are too busy to talk to.

\textbf{time\_orientation}: present

\textbf{content\_type}: current\_state

\textbf{event\_time}: unknown

\textbf{content.text}: Seeker has a few friends they think they can talk to about their feelings, but they feel the friends are too busy with work.

\textbf{content.participants}: user

\textbf{content.emotion\_or\_state}: feeling unable to reach out

\textbf{content.intention\_or\_plan}: unknown

\textbf{evidence.session}: esc150

\textbf{evidence.turn\_span}: [5, 5]

\textbf{evidence.dialogue\_time}: 2024-07-25

\textbf{status}: complete

\textbf{completeness}: complete

\vspace{3pt}
\textbf{L\_links}: [
\{ \textbf{target}: emem\_000019; \textbf{relation}: supports; \textbf{confidence}: high \},
\{ \textbf{target}: emem\_000007; \textbf{relation}: supports; \textbf{confidence}: medium \}
]

\textbf{R\_links}: [emem\_000035]
\end{minipage}
}
\caption{Example of a generated Fact-Time-Affect memory unit with temporal links.}
\label{fig:unit_case}
\end{figure}

\subsubsection{Result Case}
Table~\ref{tab:case_study} shows three difficult ES-MemEval cases: temporal reasoning, conflict detection, and user modeling. In Q1, A-Mem retrieves a plausible but wrong work-related event, while FTA-Mem locates the time-specific argument with Jack. In Q2, A-Mem misses the contradiction between painting as a helpful coping strategy and Sarah's lack of consistent time for it. In Q3, A-Mem identifies the wrong person, whereas FTA-Mem preserves the participant relation between Jimmy and George. These examples show that FTA units help the answer model use not only relevant content, but also temporal validity, status cues, and affective context.

\begin{table*}[t]
\centering
\small
\setlength{\tabcolsep}{3pt}
\renewcommand{\arraystretch}{1.08}
\begin{tabular}{p{0.27\textwidth}p{0.22\textwidth}p{0.23\textwidth}p{0.22\textwidth}}
\toprule
\textbf{Query} & \textbf{Gold Answer} & \textbf{A-Mem} & \textbf{Ours} \\
\midrule

\textbf{Q1:} As of \textbf{October 12, 2025}, what event made Sarah feel unsettled?
(\textit{Temporal Reasoning})
& An \textbf{argument with Jack} about \textbf{future plans}.
& Sarah felt unsettled due to her \textbf{boss's unexpected resignation} and concerns about leadership changes.
& Sarah felt unsettled after an \textbf{argument with Jack} about their \textbf{future plans}. \\
\midrule

\textbf{Q2:} Has Sarah found time for painting \textbf{consistently} as a coping strategy?
(\textit{Conflict Detection})
& \textbf{No}, she has found it hard to find time for painting due to \textbf{work and personal issues}.
& \textbf{Yes}, Sarah has been painting consistently as a coping strategy.
& Sarah finds painting calming, but \textbf{struggles to find time} due to \textbf{work and personal issues}. \\
\midrule

\textbf{Q3:} What future plan indicates Jimmy's willingness to \textbf{reconnect with past relationships}?
(\textit{User Modeling})
& He is considering \textbf{reaching out to George} for an honest conversation.
& Jimmy's willingness to reconnect is indicated by his desire to meet with \textbf{Jennifer}.
& Jimmy plans to \textbf{reach out to George} for an honest conversation. \\

\bottomrule
\end{tabular}
\caption{Case study on ES-MemEval. A-Mem often retrieves related but noisy memories, while FTA-Mem better identifies the target situation, temporal state, and participant relation.}
\label{tab:case_study}
\end{table*}

\subsection{Ethical Statement}

Our experiments use existing public benchmarks and do not
collect new data from real users or access private
conversations. We follow the data-use conditions of the
original benchmarks. FTA-Mem is designed to improve
evidence grounding and continuity in long-term dialogue
memory rather than to provide clinical diagnosis or
treatment.

Long-term memory may nevertheless introduce risks beyond
those of stateless dialogue systems. Incorrect, outdated, or
over-interpreted memories may persist across sessions and
influence later responses. This is particularly important for
affective information, since inferred emotional states,
intentions, and interpersonal relations may be ambiguous or
sensitive. Persistent memory may also increase privacy risks
if source dialogue spans or inferred user attributes are
retained or exposed inappropriately.

In real emotional-support applications, users should be
clearly informed when long-term memory is enabled and
should have meaningful control over consent, inspection,
correction, deletion, and data-retention settings. Deployment
would additionally require access control, secure storage,
data minimization, and safeguards for handling uncertain or
conflicting memories. We therefore view FTA-Mem as a
research framework for evaluating long-term dialogue memory
rather than a standalone clinical or counseling system.

\subsection{Prompt Templates}

We provide the core prompt templates used in FTA-Mem memory construction.
Specifically, we include the boundary-preserving window segmentation prompt,
the Fact-Time-Affect memory-unit extraction prompt, and the temporal-link
maintenance prompt. These prompts correspond to the three main construction
steps: forming coherent situation fragments, converting fragments into
structured FTA units, and maintaining relations among finalized memory units. The full prompt files and runnable configurations are included in the supplementary code. We also include the turn-level density annotation prompt used for the diagnostic information-density analysis.

\begin{figure*}[t]
\centering
\begin{minipage}{0.96\textwidth}
\begin{promptblock}{Boundary-preserving Window Segmentation Prompt}
\begin{lstlisting}
system:
You are responsible for dividing the raw dialogue of one complete session into continuous small segments suitable for memory construction.
Do not extract memory, do not answer questions. Return JSON only.

user:
Current raw session dialogue turns. Each turn includes system-generated numeric turn_index, optional original turn_id, role, timestamp, and text:
{session_dialogue}

Segmentation triggers include:
- topic_change: topic changes.
- goal_change: dialogue goal changes.
- emotion_shift: emotion or state changes noticeably.
- new_time_reference: a new time reference appears.
- new_person_or_relationship: a new person or relationship appears.
- problem_to_support_transition: the dialogue moves from problem description to support/advice.
- support_strategy_transition: support strategy changes.
- homework_or_plan_creation: homework, plan, commitment, or follow-up item appears.
- correction_or_contradiction: correction, denial, or contradiction appears.
- risk_signal: a risk signal appears.
- session_start: session starts.
- unknown: unclear boundary.

Requirements:
1. Each segment must be a continuous turn range.
2. Each segment should center on one memorable subtask, not one broad topic.
3. If one broad topic contains multiple independently memorable units, split them into separate segments.
4. Do not make a single empathy statement, confirmation, thanks, closing, or generic encouragement its own segment; merge it into the adjacent related segment.
5. Do not skip turns and do not overlap turns.
6. turn_span must use the numeric turn_index values shown in the current session; do not use the original turn_id.

Return JSON. Do not add any other top-level keys:
{
  "segments": [
    {
      "turn_span": [0, 2],
      "boundary_trigger": "session_start | topic_change | goal_change | emotion_shift | new_time_reference | new_person_or_relationship | problem_to_support_transition | support_strategy_transition | homework_or_plan_creation | correction_or_contradiction | risk_signal | unknown"
    }
  ]
}
\end{lstlisting}
\end{promptblock}
\end{minipage}
\caption{Boundary-preserving window segmentation prompt.}
\label{fig:prompt_bws}
\end{figure*}

\begin{figure*}[t]
\centering
\begin{minipage}{0.96\textwidth}
\begin{promptblock}{Fact-Time-Affect Unit Extraction Prompt}
\begin{lstlisting}
system:
You are the memory unit construction module for a long-term psychological support agent.
You must output strict JSON. Use "unknown" for missing information. Do not make unsupported inferences.

user:
Current raw dialogue turns for this segment. Each turn includes system-generated numeric turn_index, optional original turn_id, role, timestamp, and text.
segment_dialogue:{segment_dialogue}
Incomplete events left by the previous segment. If empty, there is no event that needs priority completion.
incomplete_memory:{incomplete_memory}
Follow these steps:
1. First decide whether the current segment can complete any event in incomplete_memory. If not, create new memories.
2. Construct minimal memory_units. One coherent event, state, plan, or support interaction should correspond to one unit.
3. Do not create low-value memory units for greetings, confirmations, thanks, short acknowledgements, closings, or generic assistant filler.
4. time_orientation must be one of past, present, future, mixed, unknown.
5. content_type must be exactly one of event, current_state, relationship_event, belief_candidate, homework_or_plan, support_interaction, risk_signal, general_fact.
6. status is the event state: complete if sufficiently stated, partial if meaningful but underspecified, pending for unfinished future plan/homework/follow-up, unknown if unclear.
7. evidence_turn_span use numeric turn_index values and is the continuous evidence range supporting this memory unit, not just the single trigger sentence.
8. For question-answer facts, conflicts, durations, names, yes/no facts, and relationship details, evidence_turn_span must cover the complete question-answer exchange and key follow-up turns inside the current segment.
9. content.text is the factual anchor of the memory unit. Write it as the shortest evidence-grounded factual statement that can answer future questions, close to the original wording when possible.
10. content.text must preserve QA-sensitive facts when present: names, dates, durations, quantities, negations, yes/no facts, relationship roles, causal links, commitments, plans, and explicit uncertainty. Do not replace these facts with broad psychological interpretations.
11. Use summary for a compact abstraction of the unit; do not rely on summary to carry exact facts that should be in content.text.
12. Each memory unit may output only the fields listed in the schema.

Return only the following JSON object. Do not add any other top-level keys or fields:
{
  "memory_units": [
    {
      "summary": "...",
      "time_orientation": "past | present | future | mixed | unknown",
      "content_type": "event | current_state | relationship_event | belief_candidate | homework_or_plan | support_interaction | risk_signal | general_fact",
      "event_time": "unknown",
      "content": {
        "text": "short evidence-grounded factual statement with exact entities, values, negations, or plans when present",
        "participants": "user | assistant | user,assistant",
        "emotion_or_state": "unknown",
        "intention_or_plan": "unknown"
      },
      "evidence_turn_span": [0, 3],
      "status": "complete | partial | pending | unknown"
    }
  ]
}
\end{lstlisting}
\end{promptblock}
\end{minipage}
\caption{Fact-Time-Affect memory-unit extraction prompt.}
\label{fig:prompt_fta_extract}
\end{figure*}

\begin{figure*}[t]
\centering
\begin{minipage}{0.96\textwidth}
\begin{promptblock}{Temporal Link Maintenance Prompt}
\begin{lstlisting}
system:
You maintain relations among episodic memory units.
The input contains only one newly generated memory unit and candidate neighbor memory units retrieved by similarity.
Only judge relations and provide update suggestions for each candidate neighbor memory unit's completeness, lifecycle.status, and link relation.
Return JSON only.

user:
Newly generated memory unit:
{new_memory_unit}

Candidate neighbor memory units:
{candidate_neighbor_memory_units}

Judge the relation between the new memory and each candidate neighbor:
- same_episode_as: they describe the same event, state, or plan.
- updates: the new memory adds details, progress, or completion information to an old memory.
- contradicts: the new memory clearly conflicts with an old memory.
- followup_of: the new memory is a follow-up action, result, or continuation of an old memory.
- supports: the new memory provides supporting evidence or reinforces an old memory.
- unknown: they are similar, but the specific relation cannot be confirmed.

lifecycle.status must be one of:
pending, completed, contradicted, archived, unknown.

completeness must be one of:
complete, partial, unknown.

Return one update per related candidate neighbor. Do not output updated_at; the system will set it from the new memory's event_time.
Return JSON. Do not add any other top-level keys.
{
  "memory_updates": [
    {
      "memory_id": "candidate neighbor id",
      "completeness": "complete | partial | unknown",
      "lifecycle_status": "pending | completed | contradicted | archived | unknown",
      "relation": "same_episode_as | updates | contradicts | followup_of | supports | unknown",
      "relation_confidence": "low | medium | high | unknown"
    }
  ]
}
\end{lstlisting}
\end{promptblock}
\end{minipage}
\caption{Temporal-link maintenance prompt.}
\label{fig:prompt_link}
\end{figure*}

\begin{figure*}[t]
\centering
\begin{promptblock}{Turn-level Density Annotation Prompt}
\begin{lstlisting}
system:
You are a careful annotation model. Output only valid JSON.

user:
Annotate the information density of one dialogue turn for long-term memory construction.
Use only the current turn, speaker role, and session time. Do not infer unsupported facts.

Definitions:
- factual_anchors: count distinct concrete facts, events, preferences, plans, states, or relationships explicitly expressed in this turn.
- entity_anchors: count distinct named people, places, organizations, roles, groups, or specific objects mentioned in this turn.
- temporal_anchors: count distinct temporal cues, including dates, durations, order words, past/present/future references, deadlines, or update timing.
- support_process_anchors: count distinct support-process cues, such as advice, coping actions, help-seeking, reflection, validation, planning, or response to support.
- implicitness: 1 explicit/self-contained; 2 partially context-dependent; 3 highly indirect or dependent on prior/following context.
- low_information: 1 if the turn is mostly social filler, vague acknowledgement, or too short to form a memory; otherwise 0.

Counting rules:
- Anchor fields are non-negative integer counts, not 0-5 rating scores.
- Do not cap anchor counts. If a turn contains many anchors, count all distinct anchors.
- Count only anchors grounded in this turn text; do not add information from outside the turn.
- A short turn may still contain anchors if it names a person, time, emotion, plan, or concrete fact.

Return only a JSON object with these keys:
factual_anchors, entity_anchors, temporal_anchors, support_process_anchors, implicitness, low_information.
Use non-negative integers for anchor counts, 1-3 for implicitness, and 0/1 for low_information.

Example output:
{"factual_anchors":1,"entity_anchors":1,"temporal_anchors":0,"support_process_anchors":0,"implicitness":1,"low_information":0}

Session time: {session_time}
Speaker: {speaker}
Turn text: {turn_text}
\end{lstlisting}
\end{promptblock}
\caption{Turn-level density annotation prompt.}
\label{fig:prompt_density}
\end{figure*}
% % ----------- Supplementary Content Ends Here -----------

\end{document}